\documentclass[10pt,twocolumn]{article} 
\usepackage{simpleConference}
\usepackage{times}
\usepackage{graphicx}
\usepackage{amssymb}
\usepackage{url,hyperref}
\usepackage{booktabs}
\usepackage{float}

\graphicspath{{media/}}     % organize your images and other figures under media/ folder

\begin{document}

\title{Rethinking industrial artificial intelligence: a unified foundation framework}

\author{Jay Lee, Hanqi Su\thanks{\textit{Corresponding author}} \\
Center for Industrial Artificial Intelligence, Department of Mechanical Engineering, \\
A. James Clark School of Engineering, University of Maryland, College Park, \\
Maryland, United States of America \\
\texttt{\{leejay, hanqisu\}@umd.edu}  \\
}

\maketitle
\thispagestyle{empty}

\begin{abstract}
Recent advancements in industrial artificial intelligence (AI) are reshaping the industry by driving smarter manufacturing, predictive maintenance, and intelligent decision-making. However, existing approaches often focus primarily on algorithms and models while overlooking the importance of systematically integrating domain knowledge, data, and models to develop more comprehensive and effective AI solutions. Therefore, the effective development and deployment of industrial AI require a more comprehensive and systematic approach. To address this gap, this paper reviews previous research, rethinks the role of industrial AI, and proposes a unified industrial AI foundation framework comprising three core modules: the knowledge module, data module, and model module. These modules help to extend and enhance the industrial AI methodology platform, supporting various industrial applications. In addition, a case study on rotating machinery diagnosis is presented to demonstrate the effectiveness of the proposed framework, and several future directions are highlighted for the development of the industrial AI foundation framework.
\end{abstract}

\section{Introduction}\label{1}
The rapid advancement of industrial artificial intelligence (AI) is reshaping industries worldwide.~\cite{jan2023artificial,lee2018industrial,leng2024unlocking,peres2020industrial}. Recent breakthroughs in technologies such as deep learning~\cite{jan2023artificial,lee2018industrial}, industrial internet of things (IIoT)~\cite{sisinni2018industrial,khalil2021deep}, large language models (LLMs)~\cite{lee2024unified,chang2024survey}, prognostics and health management~\cite{zio2022prognostics,su2024machine}, big data analytics~\cite{yan2017industrial,wang2022big}, and cyber-physical systems (CPS)~\cite{pivoto2021cyber,lee2015cyber} have accelerated the adoption of industrial AI, enabling industrial systems to extract actionable insights from vast amounts of industrial data and support intelligent decision-making. However, current approaches often overemphasize algorithms and models while lacking a unified framework that systematically integrates domain knowledge, data, and models. To unlock the full potential of industrial AI, a structured framework is needed – one that integrates domain knowledge, high-quality data, and intelligent AI models to address complex challenges in real-world industrial settings

Recognizing this gap, this paper proposes a unified industrial AI foundation framework composed of knowledge, data, and model modules to enhance the industrial AI methodology platform. The remainder of the paper is structured as follows: Section 2 reviews related work; Section 3 redefines the role of industrial AI; Section 4 introduces the proposed foundation framework; Section 5 provides a case study on rotating machinery diagnosis; Section 6 discusses future directions; and Section 7 concludes the paper.

\begin{table*}[!h]
\caption{Summary of existing frameworks for industrial artificial intelligence applications} \label{table:1}
\begin{center}
\resizebox{\textwidth}{!}{
\renewcommand{\arraystretch}{1.2}
\begin{tabular}{|l|l|l|}
\hline
\multicolumn{1}{|c|}{\textbf{Authors}} &
  \multicolumn{1}{|c|}{\textbf{Year}} &
  \multicolumn{1}{c|}{\textbf{Proposed industrial artificial intelligence (AI) frameworks}} \\ \hline
\multicolumn{1}{|c|}{\begin{tabular}[c]{@{}c@{}}Lee et al.~\cite{lee2018industrial}\end{tabular}} &
  {\begin{tabular}[c]{@{}c@{}}2018\end{tabular}} &
  \begin{tabular}[c]{@{}l@{}}Proposed an industrial AI ecosystem framework to integrate enabling technologies – data technology, analytic technology, platform \\ technology, and operation technology – within the Cyber-Physical System (5C) architecture. The framework systematically guides \\ AI implementation in smart manufacturing under the guidance of five key ABCDE elements: analytics technology, big data \\ technology, cloud or cyber technology, domain know-how, and evidence.\end{tabular} \\ \hline
\multicolumn{1}{|c|}{\begin{tabular}[c]{@{}c@{}}Zhang et al.~\cite{zhang2019reference}\end{tabular}} &
  {\begin{tabular}[c]{@{}c@{}}2019\end{tabular}} &
  \begin{tabular}[c]{@{}l@{}}Presented a comprehensive industrial AI reference framework consisting of seven key dimensions: Object (who), domain (where), \\ application stage (when), application requirement (why), intelligent technology (which), intelligent function (what), and solutions \\ (how); and offers a detailed overall planning for systematically integrating AI across diverse industrial scenarios.\end{tabular} \\ \hline
\multicolumn{1}{|c|}{\begin{tabular}[c]{@{}c@{}}Peres et al.~\cite{peres2020industrial}\end{tabular}} &
  {\begin{tabular}[c]{@{}c@{}}2020\end{tabular}} &
  \begin{tabular}[c]{@{}l@{}}Proposed a conceptual industrial AI framework highlighting essential enabling technologies (data, analytics, platforms, operations, \\ and human-machine interaction) while systematically identifying critical challenges, attributes, capabilities, design principles, and \\ common application domains in Industry 4.0.\end{tabular} \\ \hline
\begin{tabular}[c]{@{}l@{}}Yang et al.~\cite{yang2021intelligent}\end{tabular} &
  {\begin{tabular}[c]{@{}c@{}}2021\end{tabular}} &
  \begin{tabular}[c]{@{}l@{}}Presented a two-tier intelligent manufacturing framework designed for the process industry, integrating human-machine cooperation, \\ and intelligent autonomous control systems to achieve intelligent optimal decision-making.\end{tabular} \\ \hline
\begin{tabular}[c]{@{}l@{}}Ahmed et al.~\cite{ahmed2022artificial}\end{tabular} &
  {\begin{tabular}[c]{@{}c@{}}2022\end{tabular}} &
  \begin{tabular}[c]{@{}l@{}}Conducted a comprehensive survey on AI and explainable AI (XAI) methodologies within Industry 4.0, categorizing various XAI \\ approaches (model-specific, model-agnostic, local/global, visualization-based methods, and surrogate models), highlighting their \\ applicability, benefits, and challenges in industrial contexts.\end{tabular} \\ \hline
\begin{tabular}[c]{@{}l@{}}Jan et al.~\cite{jan2023artificial}\end{tabular} &
  {\begin{tabular}[c]{@{}c@{}}2023\end{tabular}} &
  \begin{tabular}[c]{@{}l@{}}Proposed a structured, four-stage industrial AI data pipeline (data acquisition/validation, data processing/fusion, model training/testing, \\ and model interpretation). They identified common themes, issues, and industry-specific solutions related to AI integration in \\ various sectors, providing insights into practical challenges and opportunities within Industry 4.0.\end{tabular} \\ \hline
\begin{tabular}[c]{@{}l@{}}Leng et al.~\cite{leng2024unlocking}\end{tabular} &
  {\begin{tabular}[c]{@{}c@{}}2024\end{tabular}} &
  \begin{tabular}[c]{@{}l@{}}Presented a technical reference framework structured in four layers (hardware infrastructure, computing engine, AI algorithms, \\ and industrial application layer together with related empowering technologies across layers), identifying three core opportunities \\ (collaborative intelligence, self-learning intelligence, and crowd intelligence) crucial for realizing Industry 5.0’s vision of \\ human-centric, resilient, and sustainable manufacturing.\end{tabular} \\ \hline
\begin{tabular}[c]{@{}l@{}}Lee and Su~\cite{lee2024unified}\end{tabular} &
  {\begin{tabular}[c]{@{}c@{}}2024\end{tabular}} &
  \begin{tabular}[c]{@{}l@{}}Proposed a unified industrial large knowledge model framework consisting of four systematic steps: (i) construction of a large \\ knowledge library; (ii) preparation of domain-specific instruction data; (iii) development of domain-specific knowledge large language \\ models; and (iv) establishment of intelligent domain expert machine learning systems; guided by the “6S Principle” (specialized \\ knowledge, scrutability, safety, scalability, sustainability, and systematization).\end{tabular} \\ \hline
\end{tabular}}
\end{center}
\end{table*}

\section{Related work}\label{2}
This section reviews existing studies and frameworks focusing on the integration of AI within various industrial applications. Several research works have proposed
frameworks aimed at enhancing industrial productivity, decision-making, and operations by leveraging AI capabilities. Table~\ref{table:1} provides a clear summary of these previous industrial AI frameworks, highlighting their
primary contributions. For instance, Lee et al.~\cite{lee2018industrial} introduced
an industrial AI ecosystem framework that integrates enabling technologies, such as data technology, analytic technology, platform technology, and operation technology,
within the established CPS (5C) architecture, guided by key ABCDE elements. Building on this foundation, Peres et al.~\cite{peres2020industrial} expanded the framework into a more comprehensive conceptual framework, emphasizing challenges, design principles, essential technologies, capabilities, attributes, and industrial application areas. On the other hand, Zhang et al.~\cite{zhang2019reference} developed a comprehensive reference framework structured around seven critical dimensions (object, domain, stage, requirement, technology, function, and solutions). In addition, Yang et al.~\cite{yang2021intelligent} proposed an intelligent manufacturing framework combining human-machine cooperation with autonomous intelligent control systems specifically tailored to the process industry. Ahmed et al.~\cite{ahmed2022artificial} highlighted AI and explainable AI (XAI) methodologies, categorizing various XAI methods and their applicability in Industry 4.0 contexts, while Jan et al.~\cite{jan2023artificial} structured a fourstage AI data pipeline, outlining data acquisition/validation, data processing/fusion, model training/testing, and model interpretation. Furthermore, Leng et al.~\cite{leng2024unlocking} introduced a fourlayer technical reference framework encompassing layers from hardware to industrial applications. Most recently, Lee and Su~\cite{lee2024unified} proposed a unified industrial large knowledge model (ILKM) framework emphasizing domain-specific knowledge integration through LLMs and machine learning (ML) approaches.

Although these frameworks have successfully identified important aspects of AI integration across different industrial applications, several limitations remain. These include the lack of systematic domain-specific knowledge integration, a tendency to focus on theoretical conceptualization without clear and objective guidelines, and an overemphasis on data and models without sufficient attention to systematic thinking that connects knowledge, data, and models. Motivated by these gaps, the following two sections revisit the role of industrial AI and present a unified industrial AI foundation framework aimed at addressing these critical shortcomings.

\section{Rethinking the role of industrial AI}\label{3}
The role of industrial AI can be best understood through a two-layer perspective, as shown in Figure 1, where the physical layer consists of humans, things, and systems,
while the digital layer consists of knowledge, data, and models. Based on recent emerging technologies such as IIoT~\cite{sisinni2018industrial,khalil2021deep}, digital twin~\cite{tao2018digital,lee2020integration}, CPS~\cite{pivoto2021cyber,lee2015cyber}, industrial big data analytics~\cite{yan2017industrial,wang2022big},  deep learning~\cite{jan2023artificial,lee2018industrial}, and LLMs~\cite{lee2024unified,chang2024survey}, industrial AI acts as a dynamic bridge between the physical and digital layers, continuously refining and applying AI-driven insights to improve and optimize the real-world industrial systems. More specifically, human expertise forms the basis of domain knowledge, which contains accumulated experience, research insights, and best practices from AI-driven approaches. With the rise of IoT technology, things, including machines, sensors, and other devices, generate vast amounts of data, which is an essential prerequisite for conducting industrial AI analytics, modeling, and decision-making in modern industry.
Moreover, systems such as CPS, IIoT systems, and other intelligent industrial systems use AI models to automate operations, enable intelligent decision-making, and improve real-time control.

While the two-layer perspective provides a conceptual understanding of how industrial AI bridges the physical and digital domains, real-world industrial settings present several challenges. In practice, developers and engineers face persistent difficulties in transforming heterogeneous data from IIoT devices into reliable, AI-ready formats due to inconsistent data quality, lack of standard preprocessing pipelines, and fragmented metadata management~\cite{sisinni2018industrial}. CPS, although widely discussed, often remains difficult to operationalize, as maintaining consistency between realworld operations and virtual models requires continuous data synchronization and robust model updating~\cite{pivoto2021cyber}. Despite its promise, digital twin development is challenged by the complexity of integrating multi-source data streams, real-time analytics, and simulation modeling requirements~\cite{tao2018digital}. Furthermore, the increasing availability of LLMs introduces new opportunities but also pose practical challenges in domain knowledge understanding, extraction, and interpretability for industrial use cases~\cite{lee2024unified}. Based on our observations from industry collaborations and project experiences, these complexities frequently result in ad-hoc solutions, isolated development efforts, and inefficiencies in scaling AI-driven applications. Therefore, these observations and challenges motivate the need for a structured foundation framework that systematically connects knowledge, data, and models, as presented in the following section.

\begin{figure}[htbp!]
\centering\includegraphics[width=0.6\linewidth]{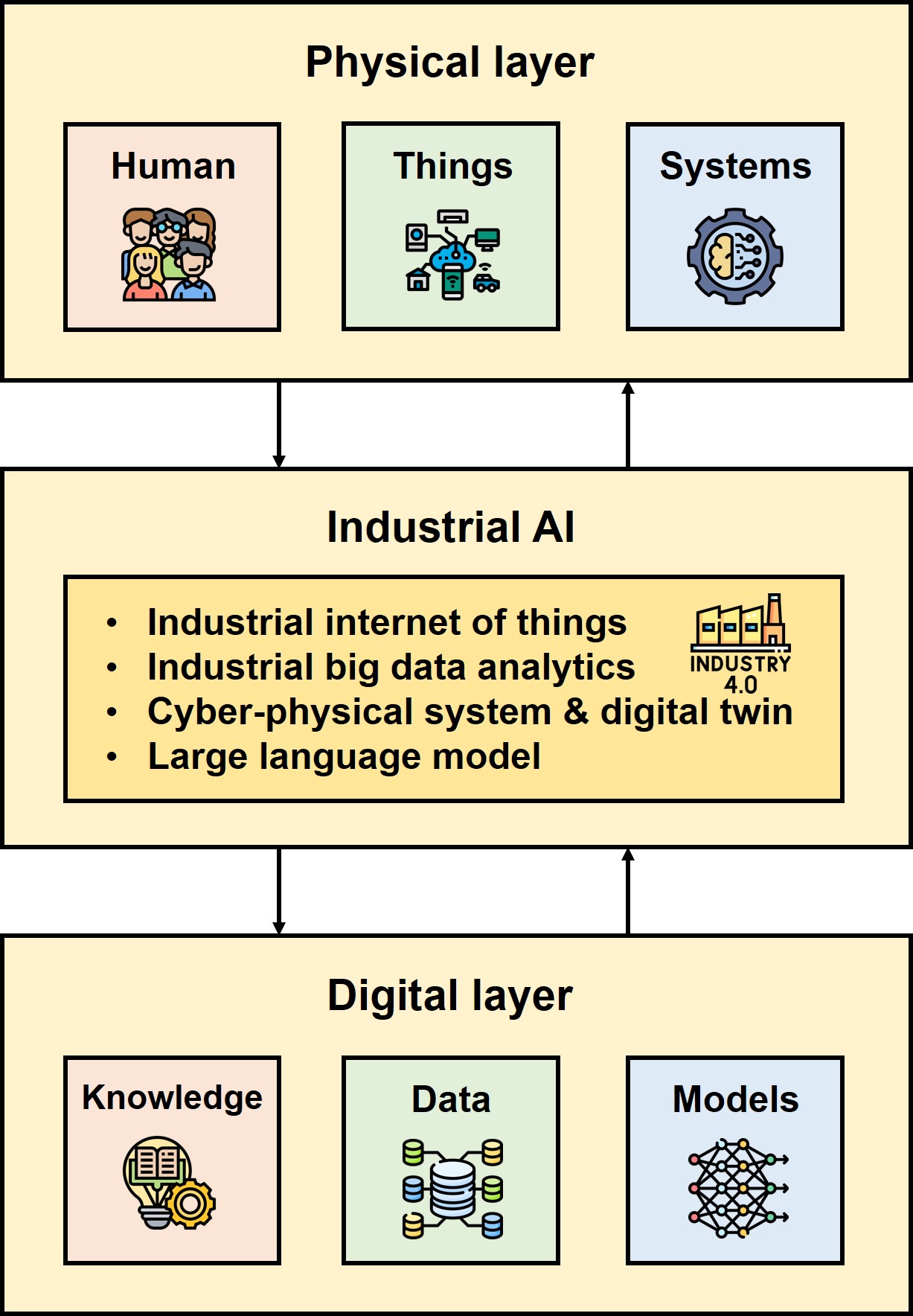}
\caption{The role of industrial artificial intelligence in bridging the physical and digital layers. Image created by the authors.}
\label{fig:1}
\end{figure}

\section{Industrial AI foundation framework}\label{4}
\begin{figure*}[h!]
\centering\includegraphics[width=\linewidth]{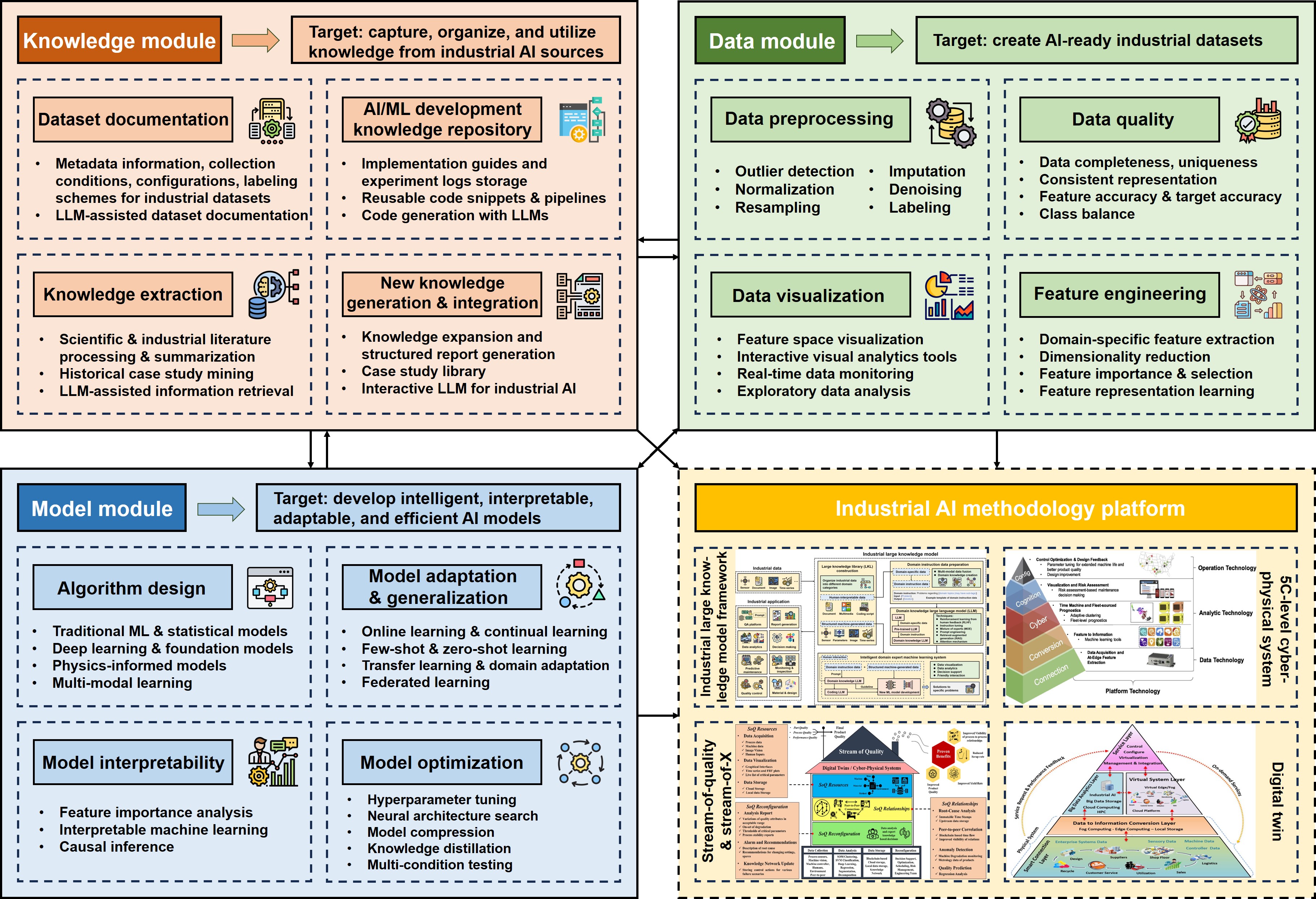}
\caption{A unified industrial artificial intelligence (AI) foundation framework. The industrial large knowledge model framework image is reprinted with permission from Lee and Su~\cite{lee2024unified}. The stream of quality image is reprinted with permission from Lee et al.~\cite{lee2022stream} Copyright \textcopyright{} 2022 Society of Manufacturing Engineers (SME). The 5C-cyber-physical system figure is reprinted with permission from Lee et al.~\cite{lee2018industrial} Copyright \textcopyright{} 2018 SME. The digital twin image is reprinted with permission from Lee et al.~\cite{lee2020integration} Copyright \textcopyright{} 2020 The Institution of Engineering and Technology. Abbreviations: LLM: Large language model; ML: Machine learning.}
\label{fig:2}
\end{figure*}

In this section, we propose a unified industrial AI foundation framework designed to systematically guide the development and deployment of industrial AI solutions, as illustrated in Figure~\ref{fig:2}. The framework consists of three core modules: (i) knowledge module; (ii) data module; and (iii) model module. Within each module, we identify four key components, resulting in 12 important aspects that collectively provide structured direction for industrial AI developers and practitioners. These modules enhance and extend the industrial AI methodology platform by providing a structured, modular foundation for existing and emerging industrial AI methodologies. Figure~\ref{fig:2} illustrates the overall structure of the proposed framework, highlighting the interconnections and dynamic feedback between knowledge, data, and model modules. Each pair of modules is connected by forward and backward arrows, highlighting that these modules continuously inform and refine one another, rather than existing in a linear sequence. In addition, forward arrows from all three modules point toward the industrial AI methodology platform, indicating that the platform is strengthened by these three modules. In Sections 4.1-4.4, the details of the knowledge, data, and model modules are outlined, followed by an explanation of how they enhance and extend the capabilities of the industrial AI methodology platform.

\subsection{Knowledge module}
The knowledge module focuses on capturing, organizing, and utilizing domain knowledge to support the entire industrial AI development cycle. In complex industrial settings, domain expertise often exists in unstructured forms, such as reports, manuals, research papers, and maintenance logs. Rather than reflecting an anthropocentric design philosophy in the traditional sense, the knowledge module aims to transform these scattered resources into structured, accessible, and reusable formats that guide data analytics and model development – serving as a foundation that enables scalable, automated, and context-aware AI system development. It focuses on the following four aspects.

\subsubsection{Knowledge extraction}
Knowledge extraction refers to the systematic process of converting unstructured information into structured, searchable knowledge representations. This is essential for enabling automated reasoning and guiding data preprocessing and model design. Techniques such as knowledge graph construction~\cite{buchgeher2021knowledge,zhong2023comprehensive,pan2024unifying} and LLM-assisted information extraction~\cite{kirk2024improving} are commonly used. For instance, tools like OpenKE~\cite{han2018openke}, TransOMCS~\cite{zhang2020transomcs}, and gBuilder~\cite{li2022gbuilder} can help with knowledge graph construction, while LLMs, such as GPT-3/4~\cite{brown2020language,achiam2023gpt}, Llama 1/2~\cite{touvron2023llama,touvron2023llama2}, PaLM~\cite{chowdhery2023palm}, and DeepSeek~\cite{liu2024deepseek}, can automate the summarization and extraction of key technical information from technical reports and research articles.

\subsubsection{Dataset documentation as industrial knowledge}
Dataset documentation is a critical part of building structured industrial knowledge. It involves systematically recording dataset metadata, collection conditions, sensor configurations, labeling schemes, and linking datasets with domain knowledge. This ensures that datasets evolve from isolated assets into long-term, reusable knowledge resources, which can be properly understood, reused, and referenced in future work. Platforms such as GitHub and Hugging Face demonstrate best practices in dataset documentation, providing clear descriptions, structured metadata fields, and version histories. In addition, LLMs could enrich dataset documentation~\cite{giner2024using}.

\subsubsection{AI/ML development knowledge repository}
An AI/ML development knowledge repository provides a centralized space to store reusable code templates, implementation guides, hyperparameter tuning records, and experiment logs. This repository accelerates development by allowing engineers to reuse proven techniques and implementation methods. Examples include maintaining shared code repositories on GitHub combined with experiment documentation platforms, using platforms like MLflow or Weights \& Biases for tracking experiments, packaging code, managing models, and sharing results. Moreover, LLMs also enhance multiple aspects, including code generation, information retrieval, and interactive AI-assisted exploration~\cite{vaithilingam2022expectation,fan2024survey,strobelt2022interactive,he2024can}.

\subsubsection{New knowledge generation and integration}
As models are developed and validated, they produce new insights and interpretations that should be continuously integrated into the existing knowledge base. This process involves capturing lessons learned, model interpretation outputs, model results, and deployment feedback, feeding them back into structured documentation and knowledge graphs. This iterative process ensures that the knowledge module remains dynamic and evolves alongside technological progress and deployment experiences.

\subsection{Data module}
Data module focuses on transforming raw industrial data into AI-ready datasets to improve data usability and reliability, enhanced by the domain knowledge from knowledge module. In processing data, this module focuses on four key aspects: data preprocessing, data quality, feature engineering, and data visualization.

\subsubsection{Data preprocessing}
This is a fundamental prerequisite for successful industrial AI applications, as raw data often contains issues such as noise, missing values, anomalies, class imbalances, and labeling inconsistencies. A well-structured data preprocessing pipeline typically involves several key techniques, including outlier detection methods (such as isolation forests, local outlier factor, and statistical thresholding); data imputation methods (including mean, median, multiple imputation, K-nearest neighbors, and regression-based approaches); signal denoising methods (such as wavelet transforms and filtering); and resampling methods (including synthetic minority over-sampling technique, random over-sampling, and under-sampling)~\cite{cofre2021big}. In addition, researchers are encouraged to adopt, design and document reusable preprocessing pipelines using standardized tools such as Pandas, scikit-learn, PyTorch, and TensorFlow to ensure reproducibility and scalability.

\subsubsection{Data quality}
Data quality directly impacts the reliability of AI models, as poor-quality data can degrade model performance and lead to poor decision-making. Key dimensions of data quality include consistent representation, completeness, uniqueness, feature accuracy, target accuracy, and target class balance~\cite{budach2022effects}. Therefore, developers should adopt systematic data quality checks as part of their pipelines. For instance, CleanML demonstrates how various data quality issues can significantly affect the performance of common machine learning models~\cite{li2021cleanml}. Furthermore, Foroni et al.~\cite{foroni2021estimating} extend conventional data quality definitions by evaluating not only how data deviates from an ideal clean dataset but also how these deviations influence task outcomes. 

\subsubsection{Feature engineering}
Feature engineering focuses on extracting meaningful representations from raw data, improving model performance, interpretability, and efficiency~\cite{zheng2018feature}. In industrial applications, well-designed features can greatly improve model accuracy. Common techniques include domain-specific feature extraction, such as time-domain statistical measures, frequency-domain features derived from Fourier transforms, and time-frequency domain features. Dimensionality reduction methods, including principal component analysis (PCA), t-distributed stochastic neighbor embedding (t-SNE), and linear discriminant analysis, help reduce complexity while maintaining important information. Feature importance ranking methods are used to identify key variables that most influence model predictions, guiding feature selection. In addition, deep learning-based feature representation learning can discover complex patterns in datasets. Tools such as tsfresh, PyCaret, and Shapley Additive Explanations (SHAP) provide automated pipelines for feature extraction, importance, and selection. Therefore, effective feature engineering ensures that AI models focus on the most informative inputs while minimizing redundancy and noise.

\subsubsection{Data visualization}
Data visualization plays an important role in making complex datasets interpretable for both humans and AI systems~\cite{allen2021data}. Visualization supports tasks such as exploratory data analysis, feature space visualization, anomaly detection, and real-time monitoring. For instance, feature space visualization using t-SNE or PCA projections can help detect outliers or clusters in high-dimensional sensor data. Real-time dashboards that visualize data streams from machines and production lines help operational teams identify issues early and improve system reliability. Moreover, industrial practitioners are encouraged to utilize interactive visualization tools like Plotly, Tableau, or D3 to enable dynamic data exploration.

\subsection{Model module}
With structured knowledge and AI-ready data in place, the model module is responsible for developing intelligent, adaptable, interpretable, and efficient AI models to drive industrial AI applications in Industry 4.0. This module provides four key aspects to help developers systematically design, adapt, and refine models that meet complex industrial requirements.

\subsubsection{Algorithm design}
Algorithm design serves as the foundation for developing AI models. Developers should select or design models that align with data characteristics, domain constraints, and application requirements. Typical options range from traditional ML and statistical methods—such as decision tree, random forest, support vector machine, gradient boosting, k-nearest neighbors, Gaussian mixture model, and various clustering algorithms—to deep learning architectures, including convolutional neural networks (CNNs), recurrent neural networks, graph neural networks, transformers, and their variants~\cite{bertolini2021machine,sarker2021deep}. These algorithms are applicable to problems involving both continuous variables (e.g., vibration signals, temperature, pressure) and discrete variables (e.g., operational states, fault modes, control events). Beyond traditional ML and deep learning architectures, physics-informed neural networks (PINNs) have emerged as valuable tools in industrial AI~\cite{xu2023physics,li2024review}. PINNs incorporate physical laws and equations into the learning process, improving model reliability in scenarios where data is sparse but prior knowledge is available. Additionally, multi-modal learning approaches, which combine information from multiple modalities — such as images, text, audio, tabular data, and sensor measurements — enable the development of more comprehensive models suitable for complex industrial scenarios~\cite{ektefaie2023multimodal,su2023multi}. Furthermore, foundation models, which are pre-trained on large-scale diverse datasets, offer a promising approach for transferable and adaptable AI across different industrial domains. These models can be fine-tuned for specific tasks, allowing for efficient deployment with reduced training time and improved generalization~\cite{zhang2025large}. In the process of algorithm design, developers may utilize prior knowledge and experience to create new model architectures or iteratively refine existing models through systematic experimentation and testing. When open-source implementations are available, they can be reproduced and further extended based on established methods.

\subsubsection{Model interpretability}
Model interpretability is crucial for ensuring that AI models are transparent and explainable. One important direction is interpretable ML, which focuses on developing models that transparent by design, such as regression models, decision trees, and generalized additive models. These models, along with techniques such as rule-based learning and sparse linear models, can provide clear and consistent explanations, making them suitable for industrial applications~\cite{linardatos2020explainable,li2022interpretable}. However, with the increasing use of deep learning models, which are often treated as black boxes, it has become challenging to understand and explain how such models make predictions. To address this, developers are encouraged to use post-hoc explanation methods such as SHAP values~\cite{lundberg2017unified}, Local Interpretable Model-agnostic Explanations~\cite{ribeiro2016should}, and integrated gradients. These methods enable feature importance analysis and provide explanations of model predictions in both global and local contexts. In addition, causal inference techniques can enhance model interpretability by identifying cause-and-effect relationships between input variables and model outcomes. Approaches such as structural causal models, counterfactual analysis, and invariant causal prediction can be used to provide more robust and stable explanations, particularly in dynamic industrial settings~\cite{li2024survey}.

\subsubsection{Model adaptation and generalization}
Model adaptation refers to the ability of an AI model to adjust its parameters or structure to accommodate new data distributions, domains, or operational scenarios without requiring complete retraining. Meanwhile, model generalization refers to the model’s ability to maintain accurate predictions when exposed to previously unseen data or tasks. They are essential to ensure that models remain robust, reliable, and effective in real-world industrial settings. Several key techniques support model adaptation and generalization. Transfer learning and domain adaptation allow models trained in one context or domain to be fine-tuned or adjusted for use in another, reducing the need for large amounts of new labeled data~\cite{singhal2023domain,li2022perspective}. Federated learning enables decentralized model training across multiple industrial sites, allowing collaborative improvement of models while preserving data privacy and security~\cite{banabilah2022federated}. In addition, few-shot and zero-shot learning techniques allow models to make accurate predictions in new domains with limited (few-shot) or no (zero-shot) task-specific training data~\cite{song2023comprehensive,wang2019survey}. These are especially valuable when collecting and labeling new industrial data is time-consuming or expensive. Online learning allows models to continuously update as new data streams become available, maintaining relevance and accuracy in real-time applications~\cite{hoi2021online}. Likewise, continual learning techniques enable models to incrementally incorporate knowledge from new data without catastrophic forgetting of previously learned information, supporting long-term model evolution in dynamic environments~\cite{wang2024comprehensive}.

\subsubsection{Model optimization and deployment readiness}
Model optimization is essential for improving the performance, efficiency, and reliability of AI models in industrial applications. It involves tuning model parameters, selecting appropriate architectures, and improving computational efficiency. Common techniques include hyperparameter optimization methods, such as grid search, random search, and Bayesian optimization, which systematically explore the parameter space to find optimal configurations~\cite{yang2020hyperparameter}. In addition to hyperparameter tuning, model optimization also addresses structural design and computational efficiency to meet industrial requirements. Neural architecture search (NAS) methods, including differentiable NAS and evolutionary algorithms, can automate the discovery of architectures that achieve optimal trade-offs between accuracy, latency, and memory consumption~\cite{elsken2019neural}. Model compression techniques such as pruning (magnitude-based or structured pruning), weight quantization, and low-rank factorization are commonly used to reduce inference time and deployment costs without significant loss of performance~\cite{deng2020model}. Knowledge distillation further enhances deployment readiness by transferring knowledge from large teacher models to smaller student models~\cite{gou2021knowledge}. Furthermore, multi-condition testing could be performed to ensure that models perform reliably across different operating scenarios~\cite{kim2022multi}.

\subsection{Industrial AI methodology platform}
The industrial AI methodology platform provides structured approaches, architectures, and guiding principles to ensure AI applications are systematically integrated into industrial systems. This platform includes multiple well-established methodologies that shape AI integration in industrial settings. Examples include: (i) ILKM: ILKM bridges LLMs with domain-specific industrial knowledge to support reasoning, explanation, and contextual decision-making in industrial AI~\cite{lee2024unified}; (ii) 5C CPS: CPS provides a structured approach for integrating AI with industrial systems through five levels: connection, conversion, cyber, cognition, and configuration, enabled by operation technology, analytic technology, data technology, and platform technology~\cite{lee2018industrial,lee2015cyber}; (iii) Stream-of-quality \& stream-of-X: This shows a paradigm that defines a structured methodology for continuous monitoring, optimization, and decision-making in industrial systems, particularly for multi-stage manufacturing processes~\cite{lee2022stream}; (iv) Digital twin: It integrates data, simulation, and services to create a virtual representation of physical entities, achieving industrial intelligence by utilizing AI, IoT, and ML techniques~\cite{lee2020integration}. 

While existing methodologies have played an important role in advancing industrial AI, their implementation can be further enhanced by the proposed industrial AI foundation framework. Rather than redefining the core methodologies, the knowledge, data, and model modules offer comprehensive and systematic views that support both existing and emerging industrial AI approaches. By promoting more structured development processes, they ensure that the design, implementation, and deployment of industrial AI solutions are more effective, reliable, and scalable. A concrete example is using ILKM to enable intelligent question-answering (QA) systems for maintenance personnel, who frequently need to query complex maintenance manuals and troubleshooting guides during equipment servicing. The knowledge module enables the structured extraction of key procedures, fault descriptions, and parameter settings from maintenance manuals and technical documentation. This content is organized into knowledge graphs and indexed repositories that can be efficiently queried by LLMs. The data module ensures that domain-specific datasets — including historical maintenance logs and annotated QA pairs — are properly documented, versioned, and linked to the knowledge base. Standardized metadata (e.g., machine type, operating conditions, and fault categories) allows the LLM to retrieve relevant answers aligned with specific equipment and scenarios. The model module guides the fine-tuning of LLMs on domain-specific QA datasets. It also defines best practices for model evaluation and optimization strategies for efficient inference. As a result, maintenance engineers can ask context-aware questions (e.g., "What should I check if vibration levels exceed the threshold in a [specific] machine?"), and the ILKM-powered QA system provides answers grounded in both structured knowledge and validated historical data.

\begin{figure*}[h!]
\centering\includegraphics[width=0.9\linewidth]{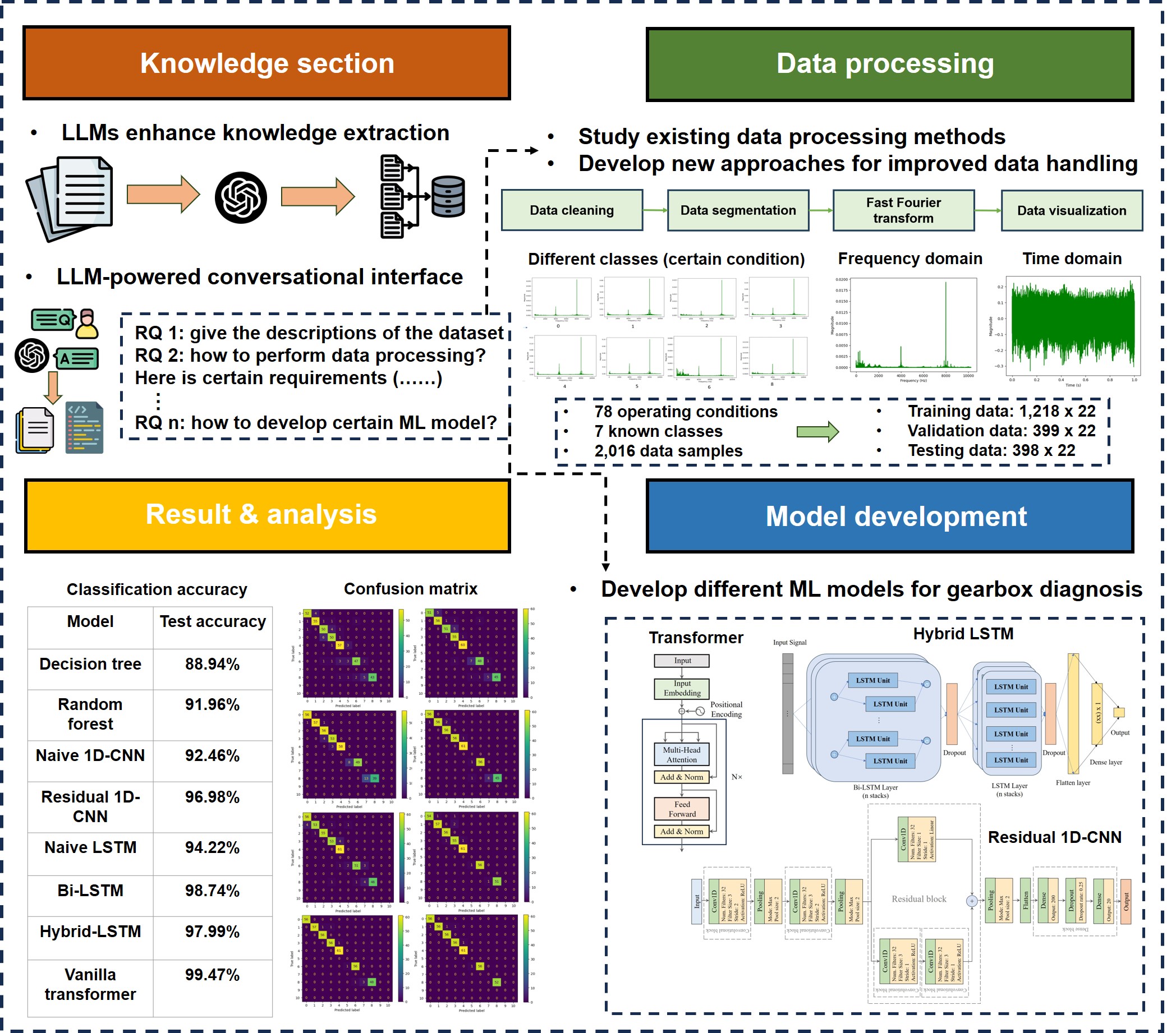}
\caption{Case study of applying the industrial artificial intelligence (AI) foundation framework for AI-driven rotating machinery fault diagnosis. Abbreviations: CNN: Convolutional neural network; LLM: Large language model; LSTM: Long short-term memory; ML: Machine learning; RQ: Research question.}
\label{fig:3}
\end{figure*}

\section{Case study}\label{5}
In this section, we demonstrate the application of the proposed industrial AI foundation framework through a case study on intelligent fault diagnosis for rotating machinery. We utilized a publicly available gearbox vibration dataset provided by the Prognostics and Health Management Society~\cite{su2024advanced}. The dataset consists of 2016 labeled samples collected under 78 different operating conditions, covering seven distinct fault classes, including various gear damage levels (class 1, 2, 3, 4, 6, and 8) and healthy state (class 0). Each sample comprises time-domain vibration signals recorded over 3 - 12 seconds at a sampling frequency of 20,480 Hz. The problem addressed in this study is evaluating the multi-class classification performance of the model using this dataset under various operating conditions. Figure~\ref{fig:3} illustrates the workflow of applying the industrial AI foundation framework, guiding knowledge extraction, data preparation, model development, and evaluation in a structured and systematic manner.

The process began with the knowledge module. Under its guidance, domain knowledge was extracted from existing publications related to the dataset by leveraging the latest LLMs, GPT-4o, via OpenAI’s application programming interface. Researchers interacted with the LLM using targeted research questions such as: “Can you provide the background information about this dataset?”; “In [target] paper, how did the authors perform data processing?”; and “How can one develop specific ML models such as 1D-CNN, LSTM, or Transformer for this task?” GPT-4o provided structured responses, summaries, and code generation examples, helping researchers quickly understand the dataset’s background, signal characteristics, data preparation requirements, feature extraction strategies, and model development practices.

The data module was applied to ensure systematic preprocessing. A structured pipeline was designed under the guidance of domain knowledge from prior studies. The process began with data cleaning to improve data quality, followed by data segmentation to increase the number of usable samples for model training. Next, feature engineering was performed using the fast Fourier transform to convert raw time-series signals into frequency domain features. The dataset was then split into training, validation, and test sets in a 3:1:1 ratio to ensure robust model evaluation. Throughout this stage, researchers continued to interact with LLMs to validate the soundness of their preprocessing strategies or to quickly obtain code examples for implementing new ideas.

After ensuring that the data were AI-ready, in the model module, researchers developed and evaluated eight different AI models based on previous methodologies and their expertise. These included: (i) tree-based model (decision tree and random forest); (ii) CNN-based model (naive 1D-CNN, residual 1D-CNN); (iii) long short-term memory (LSTM)-based model (naive LSTM, bi-LSTM, hybrid-LSTM); and (iv) transformer-based model (vanilla transformer). Deep learning models were selected for their ability to automatically extract complex patterns from large-scale, frequency-domain features and for their proven robustness in handling variations in operating conditions without the need for extensive manual feature engineering. Hyperparameter tuning and performance evaluation were performed following the experience indicated in the previous research and researchers’ development experience.

The classification accuracy and confusion matrix are presented in Figure~\ref{fig:3}, demonstrating high performance across various architectures, with the transformer model achieving the highest accuracy of 99.47\%. These results highlight the potential of advanced deep learning architectures in industrial fault diagnosis. Compared with previous studies on similar gearbox fault classification tasks, the proposed approach demonstrates clear improvement. For instance, Su and Lee~\cite{su2024advanced} developed a residual CNN that achieved 96.99\% accuracy, Vaerenberg et al.~\cite{vaerenberg2024preprocessing} used power spectral density preprocessing, log normalization, and a 3-layer CNN to reach 96.9\% accuracy, and Gauriat et al.~\cite{gauriat2024multi} proposed multi-class neural additive models with 92.03\% accuracy. The higher performance achieved in this study is attributed to not only the model design but also the systematic application of the industrial AI foundation framework. The structured guidance from the knowledge, data, and model modules ensured consistent data preprocessing, appropriate model selection, and effective hyperparameter tuning, ultimately enhancing reliability, scalability, and real-world applicability.

\section{Future direction}\label{6}
To further strengthen the industrial AI foundation framework, several key directions require further exploration. One critical aspect is talent development. Incorporating 4P-based learning (principle, practice, problem-solving, and professional) and interdisciplinary training in AI/ML, engineering, and industrial applications could benefit the next generation of industrial AI practitioners. Another promising direction is data foundry, which aims to establish a standardized framework for industrial dataset collection, annotation, benchmarking, and management. A well-structured data foundry would enhance collaborative AI research, reproducibility, and cross-industry data sharing while also facilitating the hosting of the Industrial AI Data Challenge Competitions to promote innovation and benchmark AI model performance on industrial datasets. Meanwhile, the foundation framework can be extended toward discrete event dynamic systems and hybrid control systems. This would involve adapting the knowledge, data, and model modules to better handle event-based transitions, symbolic representations, and hierarchical system logic. Furthermore, future research should explore the development of an LLM-assisted intelligent knowledge management system to better make use of historical case studies, domain expertise, and best practices. Such a system could autonomously acquire, structure, and retrieve relevant information, providing researchers and engineers with contextualized and actionable insights to improve AI model development and decision-making in industrial applications.

\section{Conclusion}\label{7}
This paper proposes a unified industrial AI foundation framework, structured into three core modules – the knowledge module, data module, and model module – which collectively support and enhance the industrial AI methodology platform. The role of industrial AI is explored in detail, and a case study on intelligent rotating machinery diagnosis demonstrates the framework’s potential. Furthermore, several future directions are discussed. Overall, our unified industrial AI foundation framework provides a systematic approach to developing, validating, and deploying industrial AI solutions for future industries. Despite these contributions, we acknowledge that acquiring and labeling industrial data remains a significant challenge. Furthermore, while the framework leverages LLMs for knowledge extraction, human guidance is still required, and computational costs may limit its deployment in certain industrial settings. Moving forward, we encourage further research and collaboration between academia, industry, and AI practitioners to refine and expand this framework, fostering the next generation of intelligent, knowledge-data-driven industrial AI ecosystems.

\bibliographystyle{abbrv}
\bibliography{manufacturing-letters}
\end{document}